\documentclass[letterpaper, 10 pt, conference]{ieeeconf}  % Comment this line out if you need a4paper

\IEEEoverridecommandlockouts                              % This command is only needed if 
\UseRawInputEncoding

\usepackage{graphics} % for pdf, bitmapped graphics files
\usepackage{epsfig} % for postscript graphics files
\usepackage{mathptmx} % assumes new font selection scheme installed
\usepackage{times} % assumes new font selection scheme installed
\usepackage{amsmath} % assumes amsmath package installed
\usepackage{amssymb}  % assumes amsmath package installed
\usepackage{multirow} 
\usepackage{gensymb}
\usepackage{booktabs}
\usepackage{balance}
\usepackage{bm}
\usepackage{makecell}
\usepackage{gensymb}
\usepackage{textcomp}
\usepackage{cite}
\makeatletter
\let\NAT@parse\undefined
\makeatother
\usepackage[colorlinks,
linkcolor=red,
anchorcolor=blue,
citecolor=black
]{hyperref}

\title{\LARGE
\textbf{TextInPlace: Indoor Visual Place Recognition in Repetitive Structures \\ with Scene Text Spotting and Verification}
}

\author{
Huaqi Tao$^{1}$,
Bingxi Liu$^{1,2}$,
Calvin Chen$^{3}$,
Tingjun Huang$^{1}$,
He Li$^{1}$,
Jinqiang Cui$^{2}$,
\\ and Hong Zhang$^{1, *}$,~\IEEEmembership{Life Fellow,~IEEE} % <-this % stops a space
\thanks{$^{1}$Southern University of Science and Technology, Shenzhen, China. 
 \texttt{taohq2024@mail.sustech.edu.cn}.
}%
\thanks{$^{2}$Peng Cheng Laboratory, Shenzhen, China.}%
\thanks{$^{3}$University of Cambridge, Cambridge, United Kingdom.} 
\thanks{*Corresponding authors: Bingxi Liu \texttt{liubx@pcl.ac.cn} and Hong Zhang \texttt{hzhang@sustech.edu.cn}.}
\thanks{This work was supported in part by the Shenzhen Science and Technology Program (No. SGDX20240115111759002), 
in part by the Meituan Academy of Robotics Shenzhen, and in part by the SUSTech High-Level Special Funds (No. G03034K003).}
}

\begin{document}

\maketitle
\thispagestyle{empty}
\pagestyle{empty}

%%%%%%%%%%%%%%%%%%%%%%%%%%%%%%%%%%%%%%%%%%%%%%%%%%%%%%%%%%%%%%%%%%%%%%%%%%%%%%%%
\begin{abstract}
Visual Place Recognition (VPR) is a crucial capability for long-term autonomous robots, enabling them to identify previously visited locations using visual information. 
However, existing methods remain limited in indoor settings due to the highly repetitive structures inherent in such environments. 
We observe that scene texts frequently appear in indoor spaces and can help distinguish visually similar but different places. 
This inspires us to propose TextInPlace, a simple yet effective VPR framework that integrates Scene Text Spotting (STS) to mitigate visual perceptual ambiguity in repetitive indoor environments. 
Specifically, TextInPlace adopts a dual-branch architecture within a local parameter sharing network. The VPR branch employs attention-based aggregation to extract global descriptors for coarse-grained retrieval, while the STS branch utilizes a bridging text spotter to detect and recognize scene texts. 
Finally, the discriminative texts are filtered to compute text similarity and re-rank the top-K retrieved images. 
To bridge the gap between current text-based repetitive indoor scene datasets and the typical scenarios encountered in robot navigation, we establish an indoor VPR benchmark dataset, called Maze-with-Text. 
Extensive experiments on both custom and public datasets demonstrate that TextInPlace achieves superior performance over existing methods that rely solely on appearance information. The dataset, code, and trained models are publicly available at \href{https://github.com/HqiTao/TextInPlace}{https://github.com/HqiTao/TextInPlace}.
\end{abstract}

%%%%%%%%%%%%%%%%%%%%%%%%%%%%%%%%%%%%%%%%%%%%%%%%%%%%%%%%%%%%%%%%%%%%%%%%%%%%%%%%

\vspace{-12pt}
\section{INTRODUCTION}

Visual Place Recognition (VPR) is a global localization task based on image retrieval, which aims to find images from a reference image database that represent the same place as the query image \cite{lowry2015visual}. It is widely applied in autonomous vehicles, augmented reality devices, and service robots. The paradigm of VPR typically involves training neural networks on large-scale datasets to encode images into compact global descriptors and retrieving the most similar images based on the similarity of these descriptors \cite{schubert2023visual}. During the past decade, researchers have conducted extensive studies to address challenges such as database scale \cite{berton2022rethinking}, viewpoint variations \cite{berton2023eigenplaces}, illumination changes \cite{liu2024npr}, seasonal transitions \cite{milford2012seqslam} and computational efficiency \cite{grainge2024design}, leading to remarkable advancements.

\begin{figure}[htbp]
    \centering
    \includegraphics[width=0.45\textwidth]{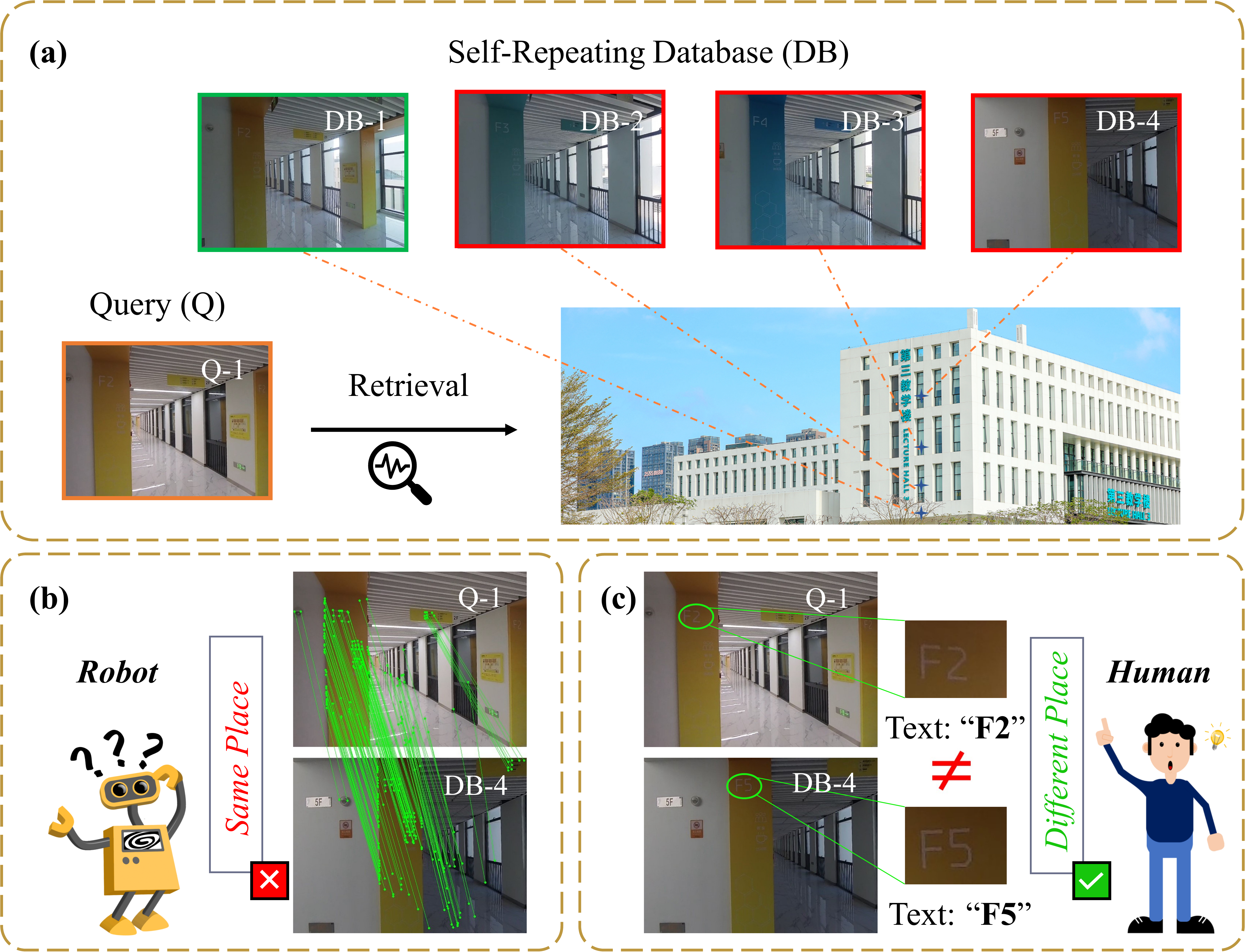}
    \caption{(a) In multi-floor buildings, there are many locations that appear highly similar but are actually different. (b) When robots rely solely on appearance information, they may mistakenly identify images of these locations as the same place. (c) In contrast, humans can utilize textual clues in the environment to accurately distinguish these highly similar places.}
    \label{1.1}
\end{figure}

However, state-of-the-art (SOTA) VPR methods often struggle in indoor scenarios, suggesting that indoor VPR remains an underexplored research area. Moreover, indoor VPR presents a more convenient and cost-effective alternative compared to traditional indoor localization methods that rely on WiFi or Radio Frequency Identification (RFID) \cite{guo2023visual}. This highlights the importance of further research and innovation to develop robust and efficient VPR techniques tailored specifically for indoor environments. We argue that the suboptimal performance of current VPR methods in indoor environments can be primarily attributed to the repeated structure caused by the repetitive and symmetrical designs of man-made buildings. Unlike outdoor environments, indoor spaces often exhibit high visual similarity across different locations (see Fig. \ref{1.1}), particularly in multi-floor buildings where architectural designs, floor layouts, and decorative elements are often highly repetitive \cite{jaenal2022unsupervised, kaveti2023challenges, taira2018inloc}. Geometry-based verification approaches are susceptible to false positive correspondences in environments with repetitive structures \cite{yu2024gv}, which significantly undermines the reliability of place recognition in unmanned systems.

\vspace{-4pt}
Inspired by human cognitive patterns, several recent studies \cite{wang2015bridging, hong2019textplace, li2023textslam, li2024resolving, jin2024robust} proposed leveraging the high-level semantic information embedded in scene texts for robot localization. 
However, these methods also typically rely on off-the-shelf Scene Text Spotting (STS) technology, which frequently becomes a computational bottleneck. 
Furthermore, they mainly use a predefined dictionary of specific words to filter out discriminative texts, and this limits their adaptability to diverse environments and requires significant manual effort.
Additionally, existing text-based localization datasets, which are collected with a focus on scene texts, remain limited in number. 
Most of these datasets deliberately capture scene texts from close distances and frontal viewpoints to facilitate text spotting. However, such data collection strategies do not accurately reflect the typical visual conditions encountered in robot navigation within real-world environments.

To enhance the applicability of VPR in indoor settings, we propose \textbf{TextInPlace}, a unique dual-branch VPR framework that harnesses textual clues for robust spatial verification in highly repetitive indoor environments. 
Specifically, TextInPlace performs coarse-grained retrieval based on the global descriptor derived from the VPR branch, and executes scene text spotting via a dedicated STS branch applied to both query and retrieved images. 
Discriminative texts are efficiently filtered out based on predefined rules or, more generally, using a large language model (LLM). 
It then establishes textual semantic correspondence by matching these discriminative scene texts, enabling explainable re-ranking of the top-K candidate images initially retrieved by VPR branch. 
Moreover, the VPR branch and STS branch are combined with a local parameter sharing backbone for optimizing computational efficiency. 
To address the limitations of existing datasets, we use a panoramic camera to establish an indoor VPR benchmark dataset, called \textbf{Maze-with-Text}. The panoramic images are then cropped to generate VPR data. In addition to scene texts, most images in the dataset also capture the structural details of the surrounding environment.

Our main contributions can be summarized as follows:

\begin{itemize}

    \item We propose TextInPlace, a unique VPR framework that generates global descriptors and performs scene text spotting through a unified local parameter sharing network, which can optimize computational efficiency. 

    \item We develop an LLM-based text filter that flexibly and generically identifies discriminative texts.

    \item We establish Maze-with-Text, an indoor VPR benchmark dataset to highlight the challenge of repetitive structures in multi-floor buildings, with scene texts appearing at various distances and perspectives, thus more closely reflecting robot navigation scenarios.
    
    \item We achieve SOTA results on both custom and public datasets, demonstrating the robustness of text-based spatial verification in repetitive indoor environments.

\end{itemize}
\vspace{-5pt}
\section{RELATED WORK}
\vspace{-2pt}
\subsection{Visual Place Recognition Network}

The VPR network consists mainly of backbone architecture and aggregation layer. Both CNN-based \cite{berton2022rethinking} \cite{berton2023eigenplaces} \cite{berton2022deep} and Transformer-based \cite{wang2022transvpr} \cite{zhu2023r2former} feature extraction backbones have been widely employed in VPR research, demonstrating their effectiveness. Recently, researchers adopted the vision foundation model DINOv2 \cite{oquab2023dinov2} as a backbone, exploring both train-free \cite{keetha2023anyloc} and fine-tuning \cite{selavpr} \cite{lu2024cricavpr} approaches, and achieving remarkable capabilities. However, vision foundation models are computationally intensive, and relying on large backbones may lead researchers to overlook innovations in other aspects of the VPR problem \cite{khaliq2025vlad}. 
The role of the aggregation layer is to compress high-dimensional feature maps into a feature vector, serving as a global descriptor. 
Early aggregation layers, including  NetVLAD \cite{arandjelovic2016netvlad}, GeM \cite{radenovic2018fine} and ConvAP \cite{ali2022gsv}, are primarily designed to be integrated with CNN-based backbone networks for enhanced feature representation. 
With the exceptional performance exhibited by the DINOv2, researchers also introduced novel aggregation techniques, such as SALAD \cite{izquierdo2024optimal}, specifically designed for networks utilizing DINOv2 as their backbone. 
More recently, BoQ \cite{ali2024boq} introduced an attention-based aggregation method, which leverages learned queries to enhance feature aggregation. 
However, most current VPR research primarily focuses on outdoor scenarios. Recent studies, such as AnyLoc \cite{keetha2023anyloc} and VLAD-BuFF \cite{khaliq2025vlad}, highlighted the poor generalization of outdoor-trained models to indoor and structured environments.

Datasets are another crucial component of learning-based VPR methods. 
Unlike outdoor VPR datasets, indoor VPR datasets are difficult to obtain from online resources and often require self-collection, resulting in relatively smaller scales. 
NYC-Indoor-VPR \cite{sheng2024nyc} introduced an indoor VPR dataset with a semi-automatic annotation method to mitigate the data scarcity in indoor VPR research. 
Additionally, existing large-scale indoor visual localization datasets, such as RISEdb \cite{sanchez2021risedb} and Naver Labs \cite{lee2021large}, can also generate VPR labels by processing the 6-degree-of-freedom (6-DOF) poses associated with the images. 

\vspace{-5pt}
\subsection{Text-based Localization and Place Recognition}
\vspace{-2pt}

In recent years, text-based localization methods have emerged as a promising solution for achieving robust localization in challenging environments. The high-level semantic features of scene texts demonstrate remarkable invariance to illumination changes and viewpoint variations. Moreover, these features offer effective discriminative capabilities in highly repetitive scenes.
TextPlace \cite{hong2019textplace} is the first to utilize scene texts for VPR in the context of topological localization. It demonstrates that scene text can effectively mitigate the challenges posed by extreme visual appearance variations in urban environments.
TextSLAM \cite{li2023textslam} is the first work that tightly integrates texts semantic information for loop closure detection (LCD) in visual SLAM.
Recently, TextLCD \cite{jin2024robust} integrated a multi-modal LCD method based on explicit scene texts into LiDAR SLAM frameworks and introduced a text-based repetitive scene dataset. 
However, these methods have two main limitations: (1) They treat scene text spotting as an independent module, leading to a decrease in system computational efficiency, and (2) most of these methods rely on a predefined dictionary to filter key texts, which significantly limits their adaptability to different environments. 
Furthermore, most existing methods depend on spatial-temporal constraints to construct text maps, whereas the VPR problem typically operates without temporal constraints.
\vspace{-5pt}
\section{Methodology}

\begin{figure*}[htbp]
    \centering
    \includegraphics[width=0.95\textwidth]{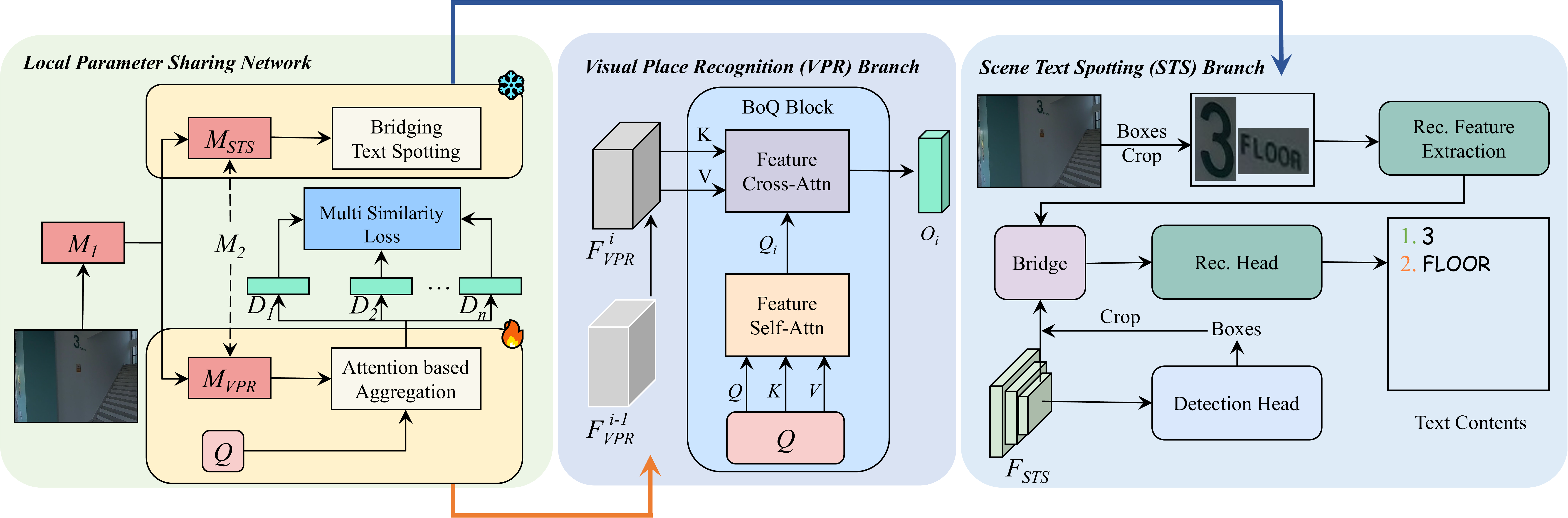}
    \caption{An overview of TextInPlace framework. TextInPlace is (a) a local parameter sharing network with two branches. (b) The Visual Place Recognition Branch extracts global descriptors for coarse global retrieval. (c) The Scene Text Spotting Branch detects and recognizes scene texts in query and database images. Finally, re-ranking is performed based on the text similarity between the query image and the retrieved images.}
    \label{fig2}
\end{figure*}

\vspace{-3pt}
In this section, we provide a detailed discussion of our proposed TextInPlace framework, including its core components and training methodology. 
The model architecture features a dual-branch design within a local parameter sharing network, as depicted in Fig. \ref{fig2}. 

\vspace{-5pt}
\subsection{The Local Parameter Sharing Network for Multi-task}

The input images $I$ are processed by a pre-trained model $M$ to obtain feature maps $F$ with dimensions $C \times H \times W$. The pre-trained model can be a classic model such as ResNet \cite{he2016deep}, a more recent model like DINOv2 \cite{oquab2023dinov2}, or other future models. We decompose $M$ into two components: $M_1$ and $M_2$. To ensure efficient inference and smooth multi-task performance, the parameters of $M_1$ are frozen during training. In contrast, $M_2$ is shared and reused across both the visual place recognition branch, denoted as $M_{\text{VPR}}$, and the scene text spotting branch, denoted as $M_{\text{STS}}$.

\vspace{-5pt}
\subsection{The Visual Place Recognition Branch}

Effective aggregation of multi-channel feature maps is crucial for accurate and fast VPR. To address this, we remove the fully connected layer in $M_\text{VPR}$ and attach an aggregator $M_\text{agg}$ based on Multi-Head Attention (MHA). This mechanism, defined as:
\begin{equation}
\text{MHA}(q, k, v) = \text{softmax} \left( \frac{q \cdot k^\top}{\sqrt{d}} \right) v,
\end{equation}
plays a key role in dynamically assessing the relevance of input features.

Specifically, we follow the aggregation block described in Bag-of-Queries (BoQ) \cite{ali2024boq} and introduce a set of learnable queries that interact with the multi-channel feature maps $F_\text{VPR}$ through cross-attention. Each block contains a fixed set of learnable queries, denoted as $Q$. The process begins by applying self-attention to $Q$:
\begin{equation}
Q_i = \text{MHA}(Q, Q, Q) + Q,
\end{equation}
where $i$ represents the block index. This step enables the learnable queries to integrate shared contextual information during the training phase.

Next, the output $Q_{i}$ is passed along with the feature maps $F_\text{VPR}^{i}$ to a cross-attention layer:
\begin{equation}
O_i = \text{MHA}(Q_i, F_\text{VPR}^{i}, F_\text{VPR}^{i}),
\end{equation}
where $F_\text{VPR}^{i}$ denotes the feature maps input to the $i$-th cross-attention layer. 
These feature maps are generated by passing $F_\text{VPR}^{i-1}$ through a transformer-encoder. 
The outputs from all blocks are then concatenated:
\begin{equation}
O = \text{Concat}(O_1, O_2, \dots, O_i),
\end{equation}
and passed through a linear projection layer to reduce the dimensionality of the final descriptor. 

These learnable queries are trainable parameters independent of the input features and dynamically aggregate information by computing attention weights between the query and feature maps.

\vspace{-5pt}
\subsection{The Scene Text Spotting Branch} 

To enable robust and efficient scene text spotting, we adopt a well-trained detector $M_{\text{det}}$ and recognizer $M_{\text{rec}}$. These modules are developed and trained independently, allowing flexible replacement and updates as needed. To further enhance their accuracy and synergy, we use an adapter $M_{\text{ada}}$ and bridge $M_{\text{bri}}$ proposed in \cite{huang2024bridging} to facilitate improved coordination between these modules. 

Given a feature map $F_{\text{STS}}$, the detection process proceeds as follows:
\begin{equation}
P_{\text{det}} = M_{\text{det}}(F_{\text{STS}}),
\end{equation}
where $P_{\text{det}}$ represents the predictions of the detector. After obtaining these predictions, we extract the corresponding regions from both the original input image $I$ and the feature map $F_{\text{STS}}$ generated by the detection backbone. This extraction process is defined as:
\begin{equation}
F_{\text{crop}} = \text{Crop}(F_{\text{STS}}, P_{\text{det}}), \quad I_{\text{crop}} = \text{Crop}(I, P_{\text{det}}),
\end{equation}
where $\text{Crop}$ denotes the crop operation based on the detector's output.

The cropped regions are then passed to the recognizer:
\begin{equation}
F_{\text{rec}} = M_{\text{rec}}(I_{\text{crop}}),
\end{equation}
where $F_{\text{rec}}$ represents the recognition features generated by the recognition backbone. Finally, both the recognition features $F_{\text{rec}}$ and the cropped features from the detection backbone $F_{\text{crop}}$ are sent to the bridge module for further processing, ensuring enhanced feature integration and consistency.

The bridge module, $M_{\text{bri}}$, is composed of a convolutional layer, a MHA layer, and a linear layer. The operation is defined as:
\begin{equation}
F = F_{\text{rec}} + \text{Linear}(\text{MHA}(\text{Conv}(F_{\text{crop}}))),
\end{equation}
where this module addresses the challenges of error accumulation and sub-optimal performance typically encountered in two-step spotting pipelines, while preserving modularity.

The adapter module, $M_{\text{ada}}$, comprises two linear layers and an activation function. It is inserted between the detector $M_{\text{det}}$ and the recognizer $M_{\text{rec}}$ to enhance the synergy between these components by refining the intermediate features and improving feature alignment between the two stages.

\vspace{-5pt}
\subsection{Verification based on Discriminative Scene Text}
Once the query descriptor $O$ is extracted, a K-nearest neighbor (KNN) search is performed in the existing feature database $D$ based on Euclidean distance. The top-K results are selected. Subsequently, the scene texts $T$ extracted from the query image is compared with the texts from the top-K results. 
It is important to note that certain text elements in scenes, such as "Fire Hydrant" or "Emergency Exit", may appear frequently across various locations, leading to potential perceptual ambiguity. To address this issue, we propose two solutions.

\subsubsection{Rule-based Text Filter} 
In indoor environments, numerical identifiers, such as door numbers, are commonly used to distinguish repetitive structures. Consequently, our rule-based text filter identifies and filters discriminative texts by detecting the presence of numerical digits within the texts.

\subsubsection{LLM-based Text Filter}
\begin{figure}[htbp]
\centering
\includegraphics[width=0.48\textwidth]{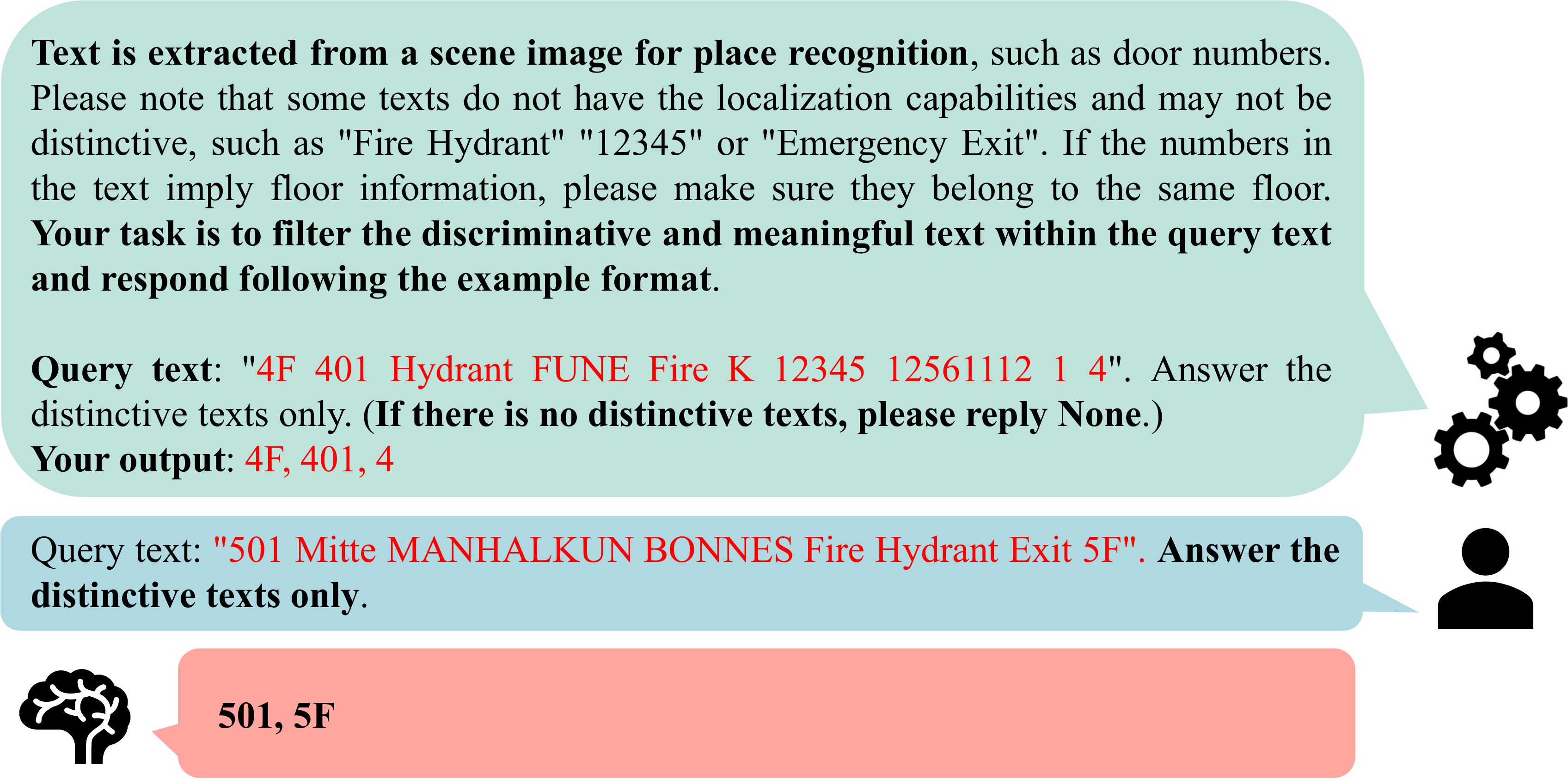}
\caption{An example prompt for the LLM-based text filter.}
\label{fig3}
\end{figure}

Due to the vast commonsense knowledge of the LLM, it can handle complex scene texts more flexibly. 
As shown in Fig. \ref{fig3}, when provided with a prompt, the LLM can automatically discern discriminative texts. 
Specifically, our implementation utilizes the GPT-4o-mini API for LLM-based text filter.

When discriminative texts are filtered, the re-ranking and verification are conducted using a string similarity score, defined as follows: 
\begin{equation}
\text{Score}_{\text{sim}}(T_{\text{query}}, T_{\text{db}}) = \frac{ |T_{\text{query}} \cap T_{\text{db}}|}{|T_{\text{query}}|},
\end{equation}
where $T_{\text{query}}$ and $T_{\text{db}}$ are the sets of detected text strings in the query image and database image, respectively. The intersection denotes matching text segments, and the score ranges between 0 and 1. The results are re-ranked based on this score, with higher scores indicating better matches.

\vspace{-5pt}
\subsection{Training Process}

In the TextInPlace framework, only the VPR branch composed of $M_{\text{VPR}}$ and $M_\text{agg}$ requires training. 
We adopt a two-phase training strategy for the VPR branch, beginning with training on a large-scale outdoor dataset and subsequently fine-tuning it using indoor data.

\subsubsection{Indoor VPR Dataset for Fine-tuning} 
\label{3.1}
Due to the limited availability of high-quality indoor VPR training datasets, we propose converting 6-DOF ground truth data from visual localization datasets into VPR labels. This approach enables the fine-tuning of outdoor-trained VPR models for indoor domains, effectively bridging the data gap and improving model performance.
Following Berton et al. \cite{berton2022rethinking}, we split the position coordinates $\{x, y\}$ into square geographical cells, and further slice each cell into a set of classes according to the direction/heading $\{heading\}$ of each image. 
Formally, the image set assigned to the class $F_{e_i, n_j, h_k}$ would be

\begin{equation}
\left\{ f : \left\lfloor \frac{\text{\textit{x}}}{M} \right\rfloor = e_i, \left\lfloor \frac{\text{\textit{y}}}{M} \right\rfloor = n_j, \left\lfloor \frac{\text{\textit{heading}}}{\alpha} \right\rfloor = h_k \right\},
\end{equation}
where $M$ and $\alpha$ are two parameters that define the spatial and angular boundaries of each class.

\subsubsection{Loss function} 
We use the multi-similarity loss \cite{wang2019multi} to train the VPR branch. For $i$-th image in training dataset, $D_i$ is the corresponding global descriptors, and $\mathcal{P}_i$ and $\mathcal{N}_j$ represent the sets of positive pairs and negative pairs, respectively. $s_{ij}$ is the cosine similarity between global descriptors of $D_i$ and $D_j$. 
The loss for samples can be computed as follows:
\begin{align}
\mathcal{L}_{\text{MS}} &= \frac{1}{N} \sum_{i=1}^{N} 
\Bigg\{ 
    \frac{1}{\alpha} \log \left[ 1 + \sum_{j \in \mathcal{P}_i} e^{-\alpha (s_{ij} - \lambda)} \right] \notag \\
    &\quad + \frac{1}{\beta} \log \left[ 1 + \sum_{j \in \mathcal{N}_i} e^{\beta (s_{ij} - \lambda)} \right] 
\Bigg\},
\end{align}
where $\alpha$, $\beta$, and $\lambda$ are hyperparameters that regulate the weighting of the positive and negative components.
\section{The Maze-with-Text Dataset}
\vspace{-2pt}
To the best of our knowledge, TextLCD \cite{jin2024robust} is the most suitable publicly available dataset for evaluating our framework, as it features repetitive structures and scene texts. 
However, the scale of the TextLCD dataset is limited, as it was collected from only three floors. Thus, it fails to capture the perceptual confusion caused by repetitive structures across multiple floors. Moreover, to simplify text spotting, a significant portion of the text images were captured at close distances and frontal viewpoints, which does not reflect the environmental structure. This also does not align with the typical scenarios encountered during robot navigation.

To address these limitations and inspired by the semi-automatic annotation method of NYC-Indoor-VPR \cite{sheng2024nyc}, we develop the Maze-with-Text dataset within a university teaching building. 
First, we use a handheld Insta360 One X3 spherical camera to record videos at $3840 \times 1920$ resolution across five floors. The higher resolution, relative to that of NYC-Indoor-VPR ($1920 \times 960$), facilitates clearer capture of scene texts for improved detection and recognition.
On each floor, two videos are captured along approximately the same trajectory at different times, serving as the query and database sets, respectively. These panoramic videos are then processed using OpenVSLAM \cite{sumikura2019openvslam} to generate a set of keyframes and topological trajectories. Finally, the turning points of the topological trajectories are manually matched to align the keyframe poses within the same coordinate frame. To avoid false positive loop closure detection, it is manually enabled only when the trajectory exhibits clear intersections.

\begin{figure}[htbp]
    \centering
    \includegraphics[width=0.45\textwidth]{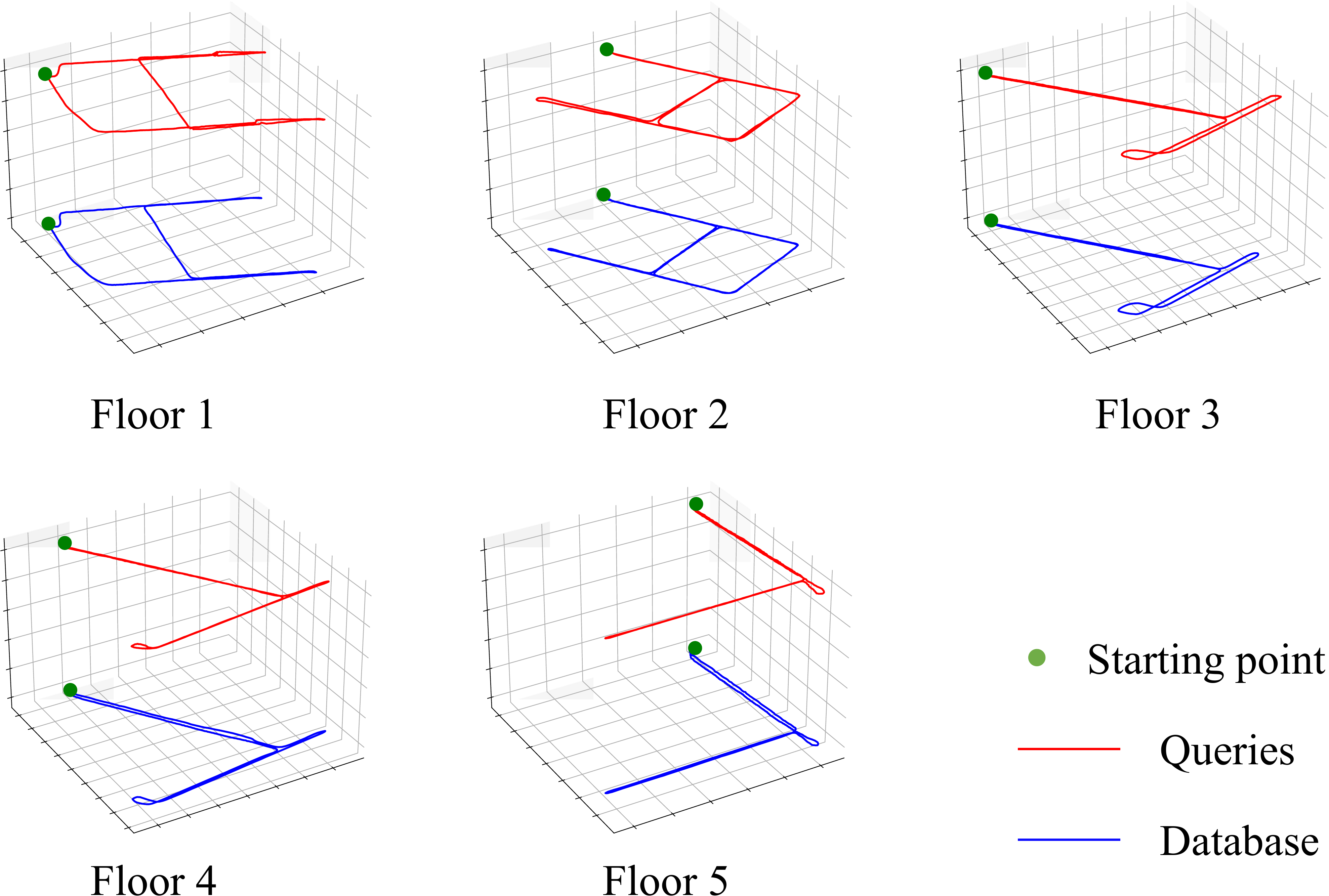}
    \caption{Illustrations of the trajectories in our dataset.}
    \label{4}
\end{figure}

The trajectories of the dataset sequences are illustrated in Fig. \ref{4}. Keyframes are used as the images in the dataset. Each equirectangular projection image is cropped into four perspective images with a resolution of $640 \times 480$ and a $90\degree$ field of view (FOV). In addition, a text detector is employed to identify and retain images containing scene texts, which are then used as query images. Table \ref{tab1} provides the number of images in the Maze-with-Text dataset.

\begin{table}[htbp]
\centering
\scriptsize 
\caption{Number of images in the Maze-with-Text dataset}
\label{tab1}
\begin{tabular}{c|cccccc} 
\hline
 & Floor 1 & Floor 2 & Floor 3 & Floor 4 & Floor 5 & All\\ \hline
Queries    & 280 & 253 & 258 & 245 & 269 & 1305   \\ \hline
Database   & 1368 & 2268 & 1588 & 1720 & 1596 & 8540   \\ \hline
\end{tabular}
\end{table}
\section{EXPERIMENTS}

\vspace{-5pt}
This section reports detailed experimental settings and evaluation results on two indoor test datasets with repetitive structures, namely Maze-with-Text and TextLCD dataset.

\vspace{-5pt}
\subsection{Implementation details}

The proposed method is implemented using PyTorch \cite{paszke2019pytorch} with model training conducted on two NVIDIA RTX 3090 GPUs, while testing is performed on a single GPU. 
We employ ResNet-50 \cite{he2016deep} backbone for feature extraction. DPText-DETR \cite{ye2023dptext} is employed for text detection, while DiG \cite{yang2022reading} is adopted for text recognition. 
Our method defaults to using a rule-based text filter, while the variant utilizing a LLM-based text filter is designated as TextInPlace-L.

We initially train the model on the GSV-Cites dataset \cite{ali2022gsv}, a large-scale urban dataset that includes 524k images and spans 62k distinct locations. 
To enable the model to generalize to indoor environments, it is fine-tuned using our reorganized Naver Labs dataset \cite{lee2021large}, as detailed in Section \ref{3.1}. Regarding hyperparameters, $M$ is set to 2, and $\alpha$ is set to 3. 
During the training phases, we use the AdamW optimizer with the learning rate set to $1 \times 10^{-4}$ for initial training and $5 \times 10^{-5}$ for fine-tuning. The batch size set to 64 places, each represented by 4 images. Moreover, training process runs for 10 epochs with an early stopping strategy, and all training images are resized to $320 \times 320$.

For evaluation metrics, we utilize the evaluation metric employed in prior research \cite{berton2022rethinking} \cite{berton2022deep} \cite{ali2022gsv} \cite{ali2024boq} \cite{ali2023mixvpr}, which entails calculating Recall@K. This metric quantifies the proportion of instances in which at least one of the top-K retrieved images falls within the predefined threshold distance of the query image. Unless otherwise specified, the test image size is set to $640 \times 480$.

\begin{table*}[htbp]
    \centering
    \renewcommand{\arraystretch}{1.2}
    \caption{Performance comparison on the Maze-with-Text dataset. \textcolor{red}{\textbf{Best}}, \textcolor{blue}{\textbf{Second Best}}, and \textbf{Third Best} results are highlighted. We report Recall@1 and Recall@5 for each method.}
    \label{tab2}
    \begin{tabular}{clccccccccc}
        \toprule
        & Method
        & Backbone
        & Dim.
        & Floor 1
        & Floor 2
        & Floor 3
        & Floor 4
        & Floor 5
        & All \\
        \midrule
        % \multirow{10}{*}{\rotatebox{90}{\textbf{1-stage}}} 
        \multirow{10}{*}{\textbf{1-stage}}
        & NetVLAD \cite{arandjelovic2016netvlad}  & VGG-16 & 32768 & 82.5 / 92.9 & 79.8 / 92.9 & 64.7 / 84.5 & 76.7 / 93.9 & 63.6 / 86.6 & 58.2 / 76.7 \\
        & SFRS \cite{ge2020self} & VGG-16 & 4096 & 86.1 / 96.4 & 89.7 / \textbf{98.4} & 71.7 / 91.5 & 84.9 / 95.1 & 56.5 / 80.7 & 59.5 / 77.5 \\
        & CosPlace \cite{berton2022rethinking} & ResNet-50  & 2048 & 85.0 / 97.1 & 82.6 / 93.3 & 53.5 / 74.0 & 81.2 / 95.9 & 37.6 / 52.8 & 50.0 / 65.6 \\
        & EigenPlaces \cite{berton2023eigenplaces} & ResNet-50 & 2048 & 88.6 / 96.8 & 90.1 / 98.0 & 71.7 / 91.9 & 85.3 / 97.1 & 56.1 / 81.0 & 60.9 / 77.5 \\
        & ConvAP \cite{ali2022gsv}  & ResNet-50 & 4096 & 85.7 / \textcolor{red}{\textbf{98.2}} & 85.0 / 94.9 & 63.6 / 82.6 & \textcolor{blue}{\textbf{88.6}} / 94.7 & 48.0 / 68.4 & 62.8 / 78.9 \\
        & MixVPR \cite{ali2023mixvpr}  & ResNet-50 & 4096 & \textbf{89.3} / \textcolor{blue}{\textbf{97.9}} & 85.8 / 98.0 & 64.3 / 83.3 & 86.9 / \textbf{98.0} & 45.0 / 63.6 & 62.4 / 79.3 \\
        & BoQ \cite{ali2024boq} & ResNet-50 & 16384 & 87.5 / 95.7 & 85.4 / 95.7 & \textbf{77.1} / \textbf{92.3} & \textbf{87.7} / 97.1 & 66.2 / 89.2 & \textbf{67.5} / \textbf{84.1} \\
        & DINOv2-BoQ \cite{ali2024boq} & DINOv2-B & 12288 & 89.3 / 96.8 & 88.5 / 98.0 & 74.0 / 89.5 & \textcolor{red}{\textbf{89.0}} / \textcolor{blue}{\textbf{98.4}} & 59.5 / 79.6 & 65.7 / 80.8 \\
        & SALAD \cite{izquierdo2024optimal} & DINOv2-B & 8448 & 87.5 / 95.4 & 83.4 / 94.9 & 67.8 / 86.8 & \textcolor{blue}{\textbf{88.6}} / \textcolor{red}{\textbf{98.8}} & 47.6 / 68.8 & 61.6 / 75.6 \\
        & CricaVPR \cite{lu2024cricavpr} & DINOv2-B & 4096 & 86.1 / \textbf{97.5} & 84.2 / 94.5 & 66.3 / 84.5 & 85.3 / 96.7 & 47.6 / 68.4 & 57.7 / 75.9 \\
        \midrule
        \multirow{5}{*}{\textbf{2-stage}}
        & Patch-NV \cite{hausler2021patch}  & VGG-16 & $2826 \times 4096$  & \textcolor{blue}{\textbf{90.7}} / 95.0 & \textbf{90.5} / 96.8 & 76.4 / 90.3 & 84.9 / 95.5 & 62.8 / 83.3 & 65.0 / 77.5 \\
        & TransVPR \cite{wang2022transvpr}  & ViT & $1200 \times 256$ & 85.4 / 95.0 & 85.8 / 94.9 & \textbf{77.1} / 91.5 & 82.4 / 92.4 & \textbf{69.1} / \textbf{90.7} & 62.9 / 78.4 \\
         & R2Former \cite{zhu2023r2former}  & ViT & $500 \times 131$ & 73.3 / 93.9 & 83.8 / 96.0 & 64.3 / 82.9 & 68.6 / 88.2 & 52.8 / 81.8 & 61.1 / 75.6 \\
        & SelaVPR \cite{selavpr}  & DINOv2-L & $61 \times 61 \times 128$ & 85.4 / 95.4 & 76.3 / 88.5 & 75.6 / 90.3 & 79.6 / 93.1 & 62.1 / 78.8 & 56.7 / 71.9 \\ 
        \cmidrule(lr){2-10}
        & TextInPlace  & ResNet-50 & 16384 & \textcolor{red}{\textbf{92.9}} / 96.8 & \textcolor{red}{\textbf{97.2}} / \textcolor{red}{\textbf{100.0}} & \textcolor{blue}{\textbf{85.3}} / \textcolor{blue}{\textbf{95.3}} & \textcolor{red}{\textbf{89.0}} / 94.7 & \textcolor{blue}{\textbf{83.6}} / \textcolor{blue}{\textbf{96.7}} & \textcolor{blue}{\textbf{85.4}} / \textcolor{blue}{\textbf{93.9}} \\
        & TextInPlace-L & ResNet-50 & 16384 & \textcolor{red}{\textbf{92.9}} / 97.1 & \textcolor{blue}{\textbf{96.4}} / \textcolor{blue}{\textbf{99.6}} & \textcolor{red}{\textbf{86.0}} / \textcolor{red}{\textbf{96.1}} & \textcolor{red}{\textbf{89.0}} / 97.6 & \textcolor{red}{\textbf{84.0}} / \textcolor{red}{\textbf{98.5}} & \textcolor{red}{\textbf{86.1}} / \textcolor{red}{\textbf{95.5}} \\
        \bottomrule
    \end{tabular}
\end{table*}

\vspace{-8pt}
\subsection{Comparison with State-of-the-art}

Table \ref{tab2} shows the the comparison of the proposed method with previous SOTA methods, including nine 1-stage models: NetVLAD \cite{arandjelovic2016netvlad},  SFRS \cite{ge2020self}, CosPlace \cite{berton2022rethinking}, EigenPlaces \cite{berton2023eigenplaces}, ConvAP \cite{ali2022gsv}, MixVPR \cite{ali2023mixvpr}, BoQ \cite{ali2024boq}, SALAD \cite{izquierdo2024optimal}, CricaVPR \cite{lu2024cricavpr}, and four 2-stage re-ranking models: Patch-NV \cite{hausler2021patch}, TransVPR \cite{wang2022transvpr}, R2Former \cite{zhu2023r2former}, SelaVPR \cite{selavpr}. 
Since some text-based methods, like TextPlace \cite{hong2019textplace}, are not open-source and mostly rely on spatial-temporal constraints (e.g., odometry) to build text maps, we do not directly compare them with our method. 
For all comparative methods, we adopted the well-trained model provided by their official sources. 
Due to the constraints on input image resolution, the test image resolution for methods employing the DINOv2 as the backbone and MixVPR is set to $224 \times 224$ and $320 \times 320$, respectively. 
All 2-stage methods re-rank the top-100 candidates to improve retrieval performance.

Compared to several recent SOTA approaches, our method demonstrates superior performance. 
TextInPlace-L exhibits a marginal performance advantage over TextInPlace, underscoring that LLM-based text filter, which dynamically adapt their filtering strategies according to the input texts, are more effective at filtering discriminative texts than static rule-based text filter.
Previous VPR methods often experience significant performance degradation due to severe perceptual confusion when encountering indoor repetitive structures. 
Notably, methods using the visual foundation model DINOv2 as the backbone network fail to significantly outperform those based on the ResNet-50 backbone network. 
2-stage methods that employ local feature-based re-ranking struggle to handle repetitive structures effectively, primarily because of the high similarity among visual elements in the scene. 
Our method addresses this challenge by leveraging scene texts as positional clues, enabling robots to distinguish such scenes in a human-like manner.

To further validate our experimental findings, we conduct comparative evaluations on a public dataset. 
Specifically, we selected sequences seq1, seq2, and seq4 from the TextLCD dataset \cite{jin2024robust}. 
The initial segments of each trajectory serves as the database, while the overlapping subsequent trajectory served as queries. 
To more closely  align with the application of loop closure detection, we select one query image every 3 seconds. It means that scene texts may not always be present in the query images.

As shown in Table \ref{tabX}, our method achieves the significant improvement on seq2, surpassing BoQ by 5.7\% in Recall@1. On seq1 and seq4, it maintains performance comparable to that of BoQ. 
Both the 2-stage method, Patch-NV, and the DINOv2-based method, SALAD, demonstrate inferior performance compared to TextInPlace. 

\setlength{\tabcolsep}{4pt}
\begin{table}[ht]
\centering
\renewcommand{\arraystretch}{1.2}
\scriptsize 
\caption{Performance comparison on TextLCD test dataset. The best is highlighted in \textbf{bold} and the second-best is \underline{underlined}.}
\label{tabX}
\begin{tabular}{cccccccc}
\toprule
\multirow{2}{*}{Method} & \multirow{2}{*}{Backbone} & \multicolumn{2}{c}{\makecell{Indoor \\ corridor (seq1)}} & \multicolumn{2}{c}{\makecell{Indoor \\ corridor (seq2)}} & \multicolumn{2}{c}{\makecell{Semi-indoor \\ corridor (seq4)}} \\ 
\cmidrule(lr){3-4} \cmidrule(lr){5-6} \cmidrule(lr){7-8} 
 & & R@1 & R@5 & R@1 & R@5 & R@1 & R@5 \\ \midrule
Patch-NV \cite{hausler2021patch} & VGG-16 & 75.7 & \underline{87.6} & 69.6 & \textbf{83.7} & \textbf{93.0} & \underline{93.0} \\
SALAD \cite{izquierdo2024optimal} & DINOv2-B & 80.4 & 85.0 & 56.0 & 66.7 & \underline{84.5} & 90.1 \\
BoQ \cite{ali2024boq} & ResNet-50 & \textbf{84.1} & \textbf{88.8} & \underline{71.6} & 77.3 & \textbf{93.0} & \textbf{94.0}  \\
TextInPlace & ResNet-50 & \underline{83.2} & \textbf{88.8} & \textbf{77.3} & \underline{81.6} & \textbf{93.0} & \underline{93.0}  \\
\bottomrule
\end{tabular}
\end{table}
\setlength{\tabcolsep}{6pt}

Since TextLCD is a multi-modal LiDAR-based LCD method, a direct comparison with our visual approach is not feasible. However, the original TextLCD paper \cite{jin2024robust} shows that its recall is lower than SALAD, which we attribute to its reliance on text-derived semantic and geometric information, neglecting appearance information. In contrast, our method incorporates global descriptors as prior information and uses discriminative texts for spatial verification, outperforming SALAD. Therefore, we argue TextInPlace theoretically outperforms TextLCD in repetitive indoor environments.

\subsection{Ablation Study}

\subsubsection{\textbf{Selection of Fine-tuning Dataset}}
In Table \ref{tab3}, we select three indoor datasets, namely NYC-Indoor \cite{sheng2024nyc}, RISEdb \cite{sanchez2021risedb}, and Naver Labs \cite{lee2021large}, for fine-tuning experiments. 
Among them, models fine-tuned on NYC-Indoor and RISEdb exhibit a decline in performance compared to their pre-fine-tuned states. 
In contrast, fine-tuning with Naver Labs leads to significant performance improvements, especially on the floor 5 subdataset.

\begin{table}[ht]
\centering
\renewcommand{\arraystretch}{1.2}
\caption{Ablation study on fine-tuning with different training set.}
\label{tab3}
\begin{tabular}{ccccccc}
\toprule
\multirow{2}{*}{ID}
& \multicolumn{3}{c}{Indoor dataset}
& \multicolumn{2}{c}{R@1} \\ 
\cmidrule(lr){2-4} \cmidrule(lr){5-6}
& NYC-Indoor & RISEdb & Naver labs & Floor 5 & All \\ 
\midrule
1&  &  &  & 59.5 & 67.7  \\
2& \checkmark &  &  & 46.1 & 59.6 \\
3&  & \checkmark&  & 55.8 & 56.0  \\
4& &  & \checkmark & \textbf{72.9} & \textbf{72.5}  \\ 
\bottomrule
\end{tabular}
\end{table}

\begin{figure*}[htbp]
    \centering
    \includegraphics[width=0.95\textwidth]{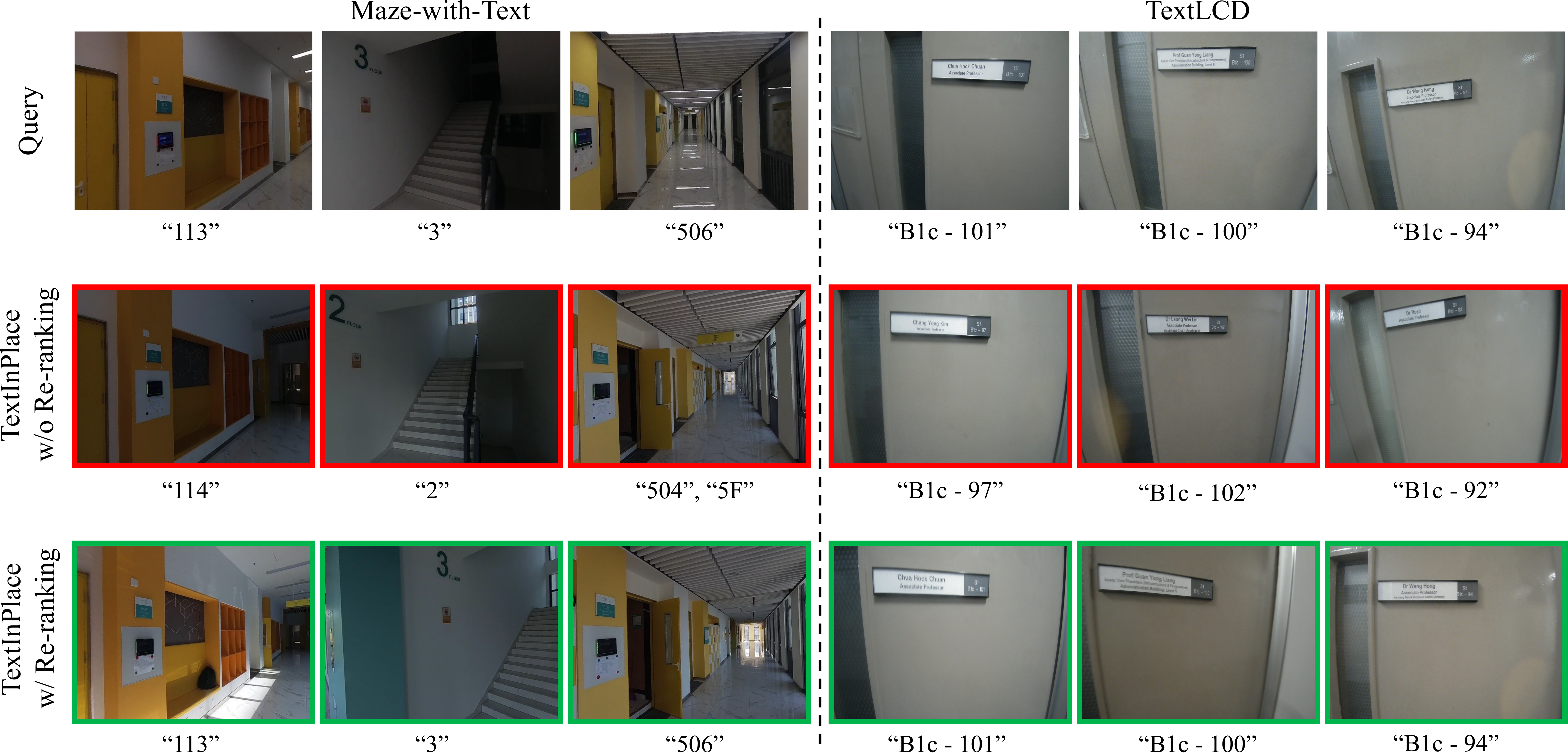}
    \caption{Examples of retrieval results for challenging queries. Each column corresponds to one query case. The first three columns are from the Maze-with-Text dataset, and the last three columns are from the TextLCD dataset. The first row displays the query image, while the other two rows show the top-1 retrieved images. Incorrect retrievals are distinguished by a red border, whereas correct retrievals are highlighted with a green border. Below each image is the discriminative scene text extracted from the image.}
    \label{7}
\end{figure*}

\subsubsection{\textbf{Effectiveness of Re-ranking and Fine-tuning}}
As shown in Table \ref{tab4}, We conduct ablation studies to evaluate the impact of re-ranking and fine-tuning in our method. Experimental results indicate that under a repetitive structure, text-based re-ranking effectively mitigates perceptual confusion caused by similar appearances. 
Indoor domain fine-tuning enhances the ability of our model to learn structured features specific to indoor environments. 
Notably, the combination of these two components leverages their individual strengths, resulting in robust and improved performance.

\begin{table}[ht]
\centering
\renewcommand{\arraystretch}{1.2}
\scriptsize 
\caption{Ablation study on re-ranking and fine-tuning.}
\label{tab4}
\begin{tabular}{ccccccc}
\toprule
\multirow{2}{*}{ID} & \multirow{2}{*}{Re-rank} & \multirow{2}{*}{Fine-tune} & \multicolumn{2}{c}{Floor 5} & \multicolumn{2}{c}{All} \\ 
\cmidrule(lr){4-5} \cmidrule(lr){6-7}
 & & & R@1 & R@5 & R@1 & R@5 \\ \midrule
1 &  &  & 59.5 & 80.7 & 67.7 & 83.8 \\
2 & \checkmark &  & 78.4\tiny{ +18.9} & 90.7\tiny{ +10.0} & 81.4\tiny{ +13.7} & 91.4\tiny{ +7.6} \\
3 &  & \checkmark & 72.9\tiny{ +13.4} & 93.7\tiny{ +13.0} & 72.5\tiny{ +4.8} & 89.4\tiny{ +5.6} \\
4 & \checkmark & \checkmark & \textbf{83.6\tiny{ +24.1}} & \textbf{96.7\tiny{ +16.0}} & \textbf{85.4\tiny{ +17.7}} & \textbf{93.9\tiny{ +10.1}} \\
\bottomrule
\end{tabular}
\end{table}

\vspace{-5pt}
\subsection{Complexity Analysis}
\vspace{-2pt}

The complexity of the network plays a vital role in practical applications. 
In Table \ref{tab5}, we compare the computational cost of our proposed method with the baseline method (BoQ + DG-Bridge Spotter) on our all floors dataset. 
In contrast to our approach, the baseline method employs BoQ \cite{ali2024boq} for image retrieval and DG-Bridge Spotter \cite{huang2024bridging} for text-based re-ranking as separate processes. 
Our method achieves notable efficiency improvements, particularly in extraction time and STS time. 
These results underscore the effectiveness of our local parameter sharing network in minimizing computational latency. 
In the future, we can replace the text detection and recognition modules with more lightweight alternatives to further reduce the time consumption of STS. 
Additionally, TextInPlace-L incurs significantly higher time consumption during the re-ranking stage compared to TextInPlace, primarily due to its reliance on remote LLM access.

\begin{table}[ht]
\centering
\renewcommand{\arraystretch}{1.2}
\caption{Comparison of computational cost.}
\label{tab5}
\begin{tabular}{ccccc}
\toprule
\multirow{2}{*}{Method} & \multicolumn{4}{c}{Latency per Query (ms)} \\ 
\cmidrule(lr){2-5}
 & Extrac. & Retrival & STS  & Re-rank \\
\midrule
BoQ + DG-Bridge Spotter & 7.7& 2.9 & 165.7 & 0.1 \\
TextInPlace & 5.9  & 2.9 & 161.0 & 0.1\\
TextInPlace-L & 5.9  & 2.9 & 161.0 & 962.8\\
\bottomrule
\end{tabular}
\end{table}

\vspace{-5pt}
\subsection{Visualization}
\vspace{-2pt}

Fig. \ref{7} shows detailed cases where repetitive structures cause significant perceptual confusion. 
When relying solely on visual descriptors, global retrieval fails in the top-1 results. 
However, by applying text-based verification, our method successfully finds the correct reference image. 
This demonstrates the effectiveness of our approach in using text-based verification to alleviate the perceptual confusion caused by repetitive structures. 
Furthermore, our dataset not only includes scene texts but also captures the environmental structure, making it a more accurate reflection of real-world robot navigation scenarios compared to TextLCD.
\vspace{-2pt}
\section{CONCLUSIONS}

% A conclusion section is not required. Although a conclusion may review the main points of the paper, do not replicate the abstract as the conclusion. A conclusion might elaborate on the importance of the work or suggest applications and extensions. 

%\addtolength{\textheight}{-12cm}   % This command serves to balance the column lengths
                                  % on the last page of the document manually. It shortens
                                  % the textheight of the last page by a suitable amount.
                                  % This command does not take effect until the next page
                                  % so it should come on the page before the last. Make
                                  % sure that you do not shorten the textheight too much.
To address the challenge of VPR in repetitive indoor environments, TextInPlace is proposed to enable coarse-grained image retrieval using global descriptors, followed by spatial verification through discriminative scene texts. 
Extensive experiments conducted on the newly proposed Maze-with-Text dataset, as well as the publicly available TextLCD dataset, demonstrate that the proposed TextInPlace can achieve state-of-the-art performance against perceptual aliasing caused by repetitive indoor structures. 
Our code and dataset are available for the benefit of the community.

%%%%%%%%%%%%%%%%%%%%%%%%%%%%%%%%%%%%%%%%%%%%%%%%%%%%%%%%%%%%%%%%%%%%%%%%%%%%%%%%

%%%%%%%%%%%%%%%%%%%%%%%%%%%%%%%%%%%%%%%%%%%%%%%%%%%%%%%%%%%%%%%%%%%%%%%%%%%%%%%%

%%%%%%%%%%%%%%%%%%%%%%%%%%%%%%%%%%%%%%%%%%%%%%%%%%%%%%%%%%%%%%%%%%%%%%%%%%%%%%%%
% \section*{APPENDIX}

% Appendixes should appear before the acknowledgment.
% \vspace{-2pt}
\section*{ACKNOWLEDGMENT}
The authors would like to thank Tongxing Jin, Thien-Minh Nguyen, Xinhang Xu, Yizhuo Yang, Shenghai Yuan, Jianping Li and Lihua Xie for the contribution of the TextLCD dataset \cite{jin2024robust} and for allowing its academic use.

% The preferred spelling of the word ÒacknowledgmentÓ in America is without an ÒeÓ after the ÒgÓ. Avoid the stilted expression, ÒOne of us (R. B. G.) thanks . . .Ó  Instead, try ÒR. B. G. thanksÓ. Put sponsor acknowledgments in the unnumbered footnote on the first page.

%%%%%%%%%%%%%%%%%%%%%%%%%%%%%%%%%%%%%%%%%%%%%%%%%%%%%%%%%%%%%%%%%%%%%%%%%%%%%%%%

% References are important to the reader; therefore, each citation must be complete and correct. If at all possible, references should be commonly available publications.

\newpage
\bibliographystyle{IEEEtran}
\bibliography{references}

\end{document}